\pdfoutput=1

\documentclass[11pt]{article}

\usepackage{acl}

\usepackage{times}
\usepackage{latexsym}
\usepackage{booktabs}
\usepackage{authblk}

\usepackage[T1]{fontenc}

\usepackage[utf8]{inputenc}

\usepackage{microtype}

\usepackage{inconsolata}
\usepackage{graphicx}
\usepackage{caption}
\usepackage{hhline}
\usepackage{multirow}
\usepackage{array}
\usepackage{tabularx}
\usepackage{graphicx}
\usepackage{floatrow}
\usepackage{caption}
\usepackage{amsmath}

\usepackage{float}

\title{Error Reflection Prompting: Can Large Language Models Successfully Understand Errors?}

\author{
    \textbf{Jason Li} \quad
    \textbf{Lauren Yraola} \quad
    \textbf{Kevin Zhu} \quad
    \textbf{Sean O'Brien}
}
    
\affil{Algoverse AI Research}
\affil{\texttt{jl255788@gmail.com, kevin@algoverse.us}}

\begin{document}
\maketitle
\begin{abstract}
Prompting methods for language models, such as Chain-of-thought (CoT), present intuitive step-by-step processes for problem solving. These methodologies aim to equip models with a better understanding of the correct procedures for addressing a given task. Despite these advancements, CoT lacks the ability of reflection and error correction, potentially causing a model to perpetuate mistakes and errors. Therefore, inspired by the human ability for said tasks, we propose Error Reflection Prompting (ERP) to further enhance reasoning in language models. Building upon CoT, ERP is a method comprised of an incorrect answer, error recognition, and a correct answer. This process enables the model to recognize types of errors and the steps that lead to incorrect answers, allowing the model to better discern which steps to avoid and which to take. The model is able to generate the error outlines itself with automated ERP generation, allowing for error recognition and correction to be integrated into the reasoning chain and produce scalability and reliability in the process. The results demonstrate that ERP serves as a versatile supplement to conventional CoT, ultimately contributing to more robust and capable reasoning abilities along with increased interpretability in how models ultimately reach their errors.
\end{abstract}

\section{Introduction}

Large Language Models (LLMs) have demonstrated remarkable performance across numerous NLP tasks \cite{openai2023gpt4}. Yet scaling them is costly, prompting interest in more efficient methods like Chain-of-Thought (CoT) \cite{wei2023chainofthought}, which guides step-by-step reasoning. A key shortcoming of CoT, however, is that it lacks a mechanism to recognize and correct its own mistakes \cite{turpin2023language}. This can lead models to persist with errors, as they lack explicit guidance to reflect on and revise flawed reasoning.

\begin{figure} [H]
    \includegraphics[width=3 in]{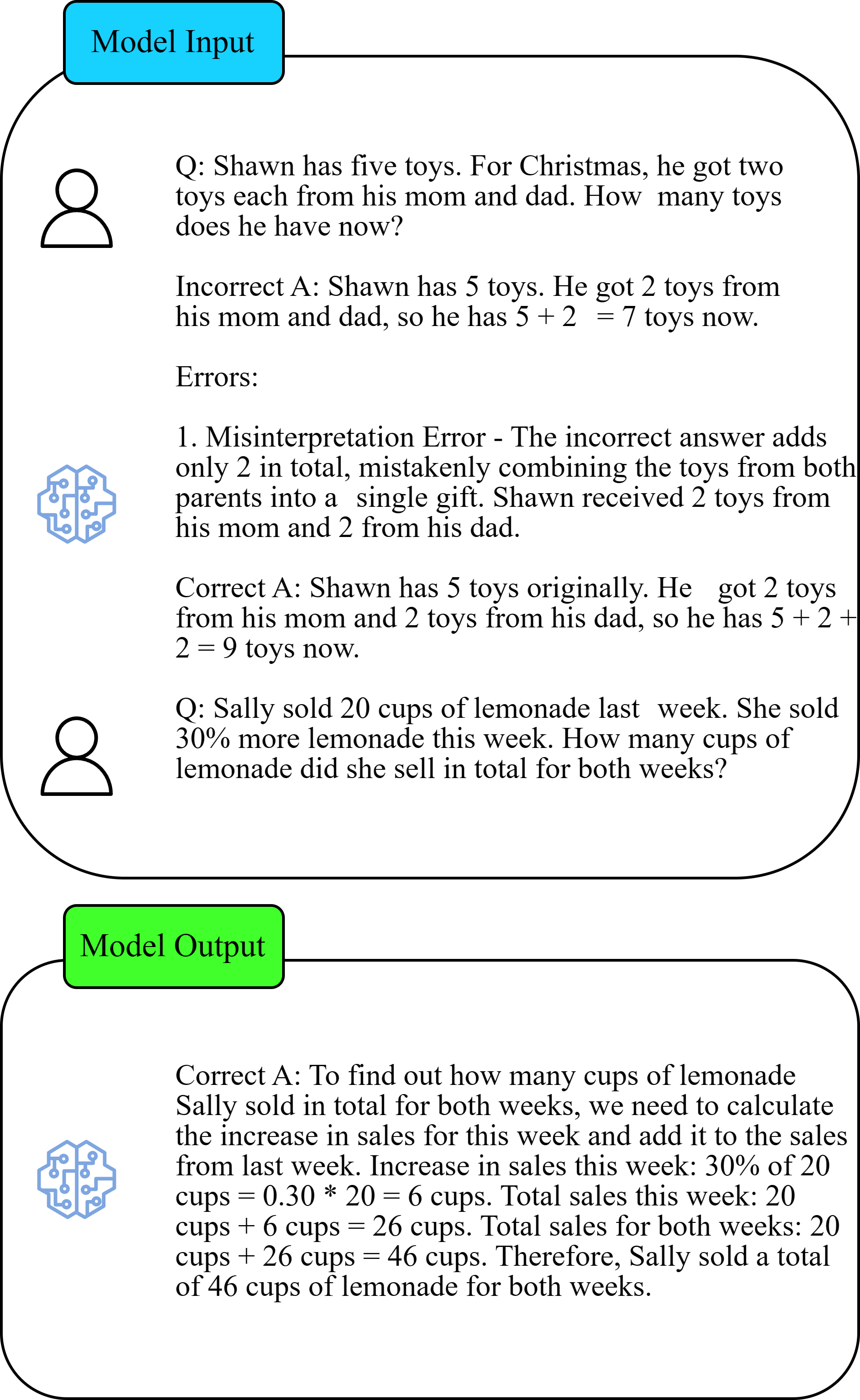}
    \caption{Example of Error Reflection Prompting with a focus on misinterpretation.}
\end{figure}

Drawing on the human ability to self-reflect and amend mistakes \cite{huang2023reasoning}, we propose Error Reflection Prompting (ERP). ERP supplies an incorrect answer alongside an explanation of the errors, then presents the correct reasoning. By exposing common pitfalls and clarifying why they are wrong, ERP steers models away from these errors in future reasoning. Evaluations on multiple benchmarks with GPT-3.5 and GPT-4 confirm that ERP consistently outperforms standard CoT, notably lowering specific error types (e.g., algebraic and calculation mistakes) and balancing the overall error distribution. Thus, ERP adds a complementary layer of self-correction to CoT, enhancing both accuracy and transparency in model outputs.

\begin{table}[ht]
  \centering
  \begin{tabular}{c c c} \hline 
    
    \textbf{Dataset}& \textbf{Type of Reasoning}& \textbf{Test}\\ \hline 
    
    GSM8K& Arithmetic& 1319\\ 
    
    AQuA& Arithmetic& 254\\ 
    
    MATH& Arithmetic& 1000\\  

    CSQA& Commonsense& 1221\\
    
 StrategyQA& Commonsense&2290\\ \hline
  \end{tabular}
  \caption{Problems used from each dataset. Note that CSQA and StrategyQA's test set answers are not publicly available, so we follow \cite{wei2023chainofthought} and evaluate performance on development (dev) datasets.}
  \label{tab:1}
\end{table}

We propose \textbf{Error Reflection Prompting (ERP)}, which prompts an LLM with an incorrect answer, explicit errors, and a corrected solution. This strategy helps the model identify and avoid known pitfalls:

\begin{enumerate} \item \textbf{Incorrect Answer.} Provide a solution with specific errors—ones the model commonly makes \cite{gao2022automated}—ensuring the prompt focuses on these weaknesses. \item \textbf{Error Reflection.} Explain each error and why it is incorrect, offering rationales to help the model understand and avoid repeating them. \item \textbf{Correct Answer.} Present the corrected chain of reasoning, reinforcing the proper approach. \end{enumerate}

Formally, let $E$ be a set of errors identified during analysis, and $Q \subseteq E$ be a subset of $n$ errors. For each example, we embed $Q$ into the prompt and provide corresponding explanations and the correct solution.

In contrast to a single, general CoT prompt, ERP may need unique prompts for each problem, which can be time-consuming. To make ERP scalable, we automate the process:

\begin{enumerate} \item Prompt the model to generate $n$ plausible errors for a given question. \item Construct an incorrect solution incorporating those errors. \item Incorporate the question, incorrect solution, errors, and correct solution into the final prompt. \end{enumerate}

By letting the model generate its own error outlines, ERP integrates recognition and correction into the reasoning chain, making it more practical for large-scale tasks.

\section{Experiments}

\subsection{Experimental Setup}
We focus our analysis on two types of reasoning: commonsense and arithmetic. By using random sampling, we ran tests on all shuffled datasets. 
\begin{table*}[ht]
  \centering
  \renewcommand{\arraystretch}{1.2}
  \begin{tabular}{l c c l}
 & & &\\
    \hline
    \multirow{2}{*}{\textbf{Prompting Method}}&\multicolumn{3}{c}{\textbf{Arithmetic Reasoning}}\\
    \hhline{~---}
    & \textbf{GSM8K Accuracy}& \textbf{AQuA Accuracy} &\textbf{MATH Accuracy}\\
    \hline
    GPT-3.5 with 5-shot Auto ERP& \textbf{79.8}\%& 48.0\%&\ \ \ \ \ \ \ \ \ --------\\
    GPT-3.5 with 4-shot CoT& 74.6\%& 54.3\%&\ \ \ \ \ \ \ \ \ 40.2\%\\
    GPT-3.5 with 4-shot ERP& 77.8\%& \textbf{58.7\%}&\ \ \ \ \ \ \ \ \ \textbf{41.1\%}\\\hline
 GPT-4 with 4-shot CoT& 95.4\%& 75.9\%&\ \ \ \ \ \ \ \ \ 54.1\%\\
 GPT-4 with 4-shot ERP& \textbf{95.7\%}& \textbf{76.7\%}&\ \ \ \ \ \ \ \ \ \textbf{54.8\%}\\
    \hhline{----}
  \end{tabular}
  \caption{Results from a variety of math-based datasets}
  \label{tab:3}
\end{table*}

\begin{table}[ht]
  \centering
  \renewcommand{\arraystretch}{1.2}
  \resizebox{1\textwidth}{!}{%
  \begin{tabular}{l c c}
    \hline
    \multirow{2}{*}{\textbf{Prompting Method}} & \multicolumn{2}{c}{\textbf{Commonsense Reasoning}} \\
    \cline{2-3}
     & \textbf{StrategyQA Acc.} & \textbf{CSQA Acc.} \\
    \hline
    GPT-3.5 with 4-shot CoT & 65.3\% & 75.8\% \\
    GPT-3.5 with 4-shot ERP & \textbf{66.1\%} & \textbf{77.3\%} \\
    GPT-3.5 with 4-shot ERP w/ Categories & 62.9\% & ------ \\\hline
    GPT-4 with 4-shot CoT & 79.9\% & 85.3\% \\
    GPT-4 with 4-shot ERP & \textbf{80.0\%} & \textbf{86.3\%} \\
    GPT-4 with 4-shot ERP w/ Categories & 76.0\% & ------ \\
    \hline
  \end{tabular}%
    }
  \caption{Results from commonsense-based datasets}
  \label{tab:4}
\end{table}
Dataset details are included in Table \ref{tab:1}.

\textbf{Arithmetic Dataset}  \indent For arithmetic reasoning, we consider three of the following datasets: 
GSM8K, a dataset for grade school math word problems \cite{DBLP:journals/corr/abs-2110-14168}, AQuA, a dataset consisting of algebraic word problems \cite{ling-etal-2017-program}, and MATH, a dataset with challenging competition math problems  \cite{hendrycks2021measuring}.  All datasets have been run on their respective test sets except for MATH, for which we have limited to 1000 test samples for a reasonable computing limit.

\textbf{Commonsense Datasets}\indent In regards to commonsense reasoning, we conduct tests on StrategyQA, which requires models to infer solutions to questions with implicit reasoning steps  \cite{geva2021did}, and CSQA, a dataset for commonsense question answering \cite{talmor2019commonsenseqa}.   

\textbf{Models} \indent We use the OpenAI Chat Completions API, one of the most widely known and used frameworks for accessing language models \cite{ge2023openagi}. We use GPT-3.5 and GPT-4 for all tests. Specifically, we use gpt-3.5-turbo-0613 and gpt-4-1106-preview respectively for all tests. In addition, we employ 4-shot prompts for each dataset, and each prompt is derived from previous works in \cite{wei2023chainofthought}.  All errors were created from human annotators or generated using GPT-4 with some slight modifications. Full prompts can be found in Appendix \ref{app:Appendix A}.

\subsection{Analysis Evaluation}
After collecting the results, we analyze the distribution of errors according to different categories. To accomplish our analysis of the model's errors, GPT-4 was used as an annotator. A sample of 100 errors from each dataset was pulled to evaluate the effectiveness of GPT-4 as an error detector. To categorize each error and understand the underlying distribution, we separated errors into five and four different classes for arithmetic and commonsense reasoning datasets respectively. For the arithmetic datasets, we have \textbf{Misinterpretation of Question}, \textbf{Missed Steps}, \textbf{Logical/Commonsense Error}, \textbf{Wrong Selection of Answer}, and \textbf{Algebraic/Calculation Error}. For more details on each of these categories, please see Table~\ref{app:error-types-table} in the Appendix. As for the commonsense datasets, we have taken a majority of our errors from the categories presented in \cite{dou2022gpt3} being \textbf{Commonsense Error}, \textbf{Encyclopedic Error}, and \textbf{Self-Contradiction Error}. We have also added an \textbf{Assumption Error} to better align with problem solving tasks. We have chosen to analyze on all datasets except for the MATH dataset \cite{hendrycks2021measuring} as the model may struggle to identify errors due to the complexity of the problems. All of these errors and their examples are displayed in Table \ref{tab:2}.

\section{Results}

\subsection{Arithmetic Results}

\textbf{ERP demonstrates increases over conventional CoT.}\indent Table \ref{tab:3} shows the differences between the two prompting methods. The largest gain is in AQuA, outperforming CoT by  \textbf{+4.4\%}. However, with the automatic prompt generation, ERP is able to outperform CoT by \textbf{+5.2\%}. Though ERP outperforms CoT on all three datasets, the MATH \cite{hendrycks2021measuring} dataset had the lowest percentage increase. It is the most difficult dataset in terms of problem solving ability, and a strong reasoning ability is required to score well on it.
\begin{figure*}[t] 
    \begin{center}
        \includegraphics[width=0.8\textwidth]{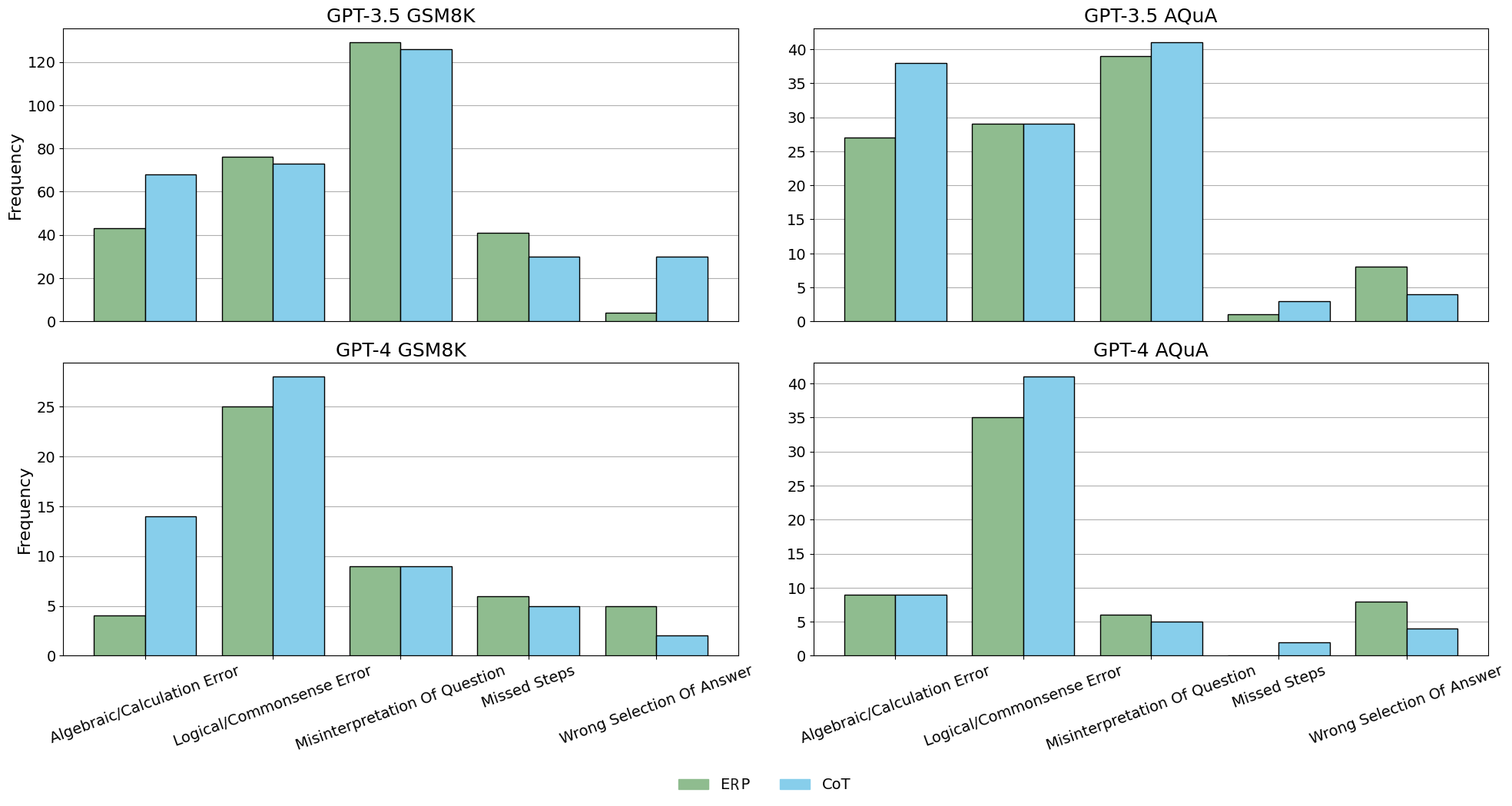} 
    \end{center}
    \caption{Error annotations on arithmetic reasoning datasets.}
\label{fig:3}
\end{figure*}

\begin{figure}[ht]
    \includegraphics[width=3in]{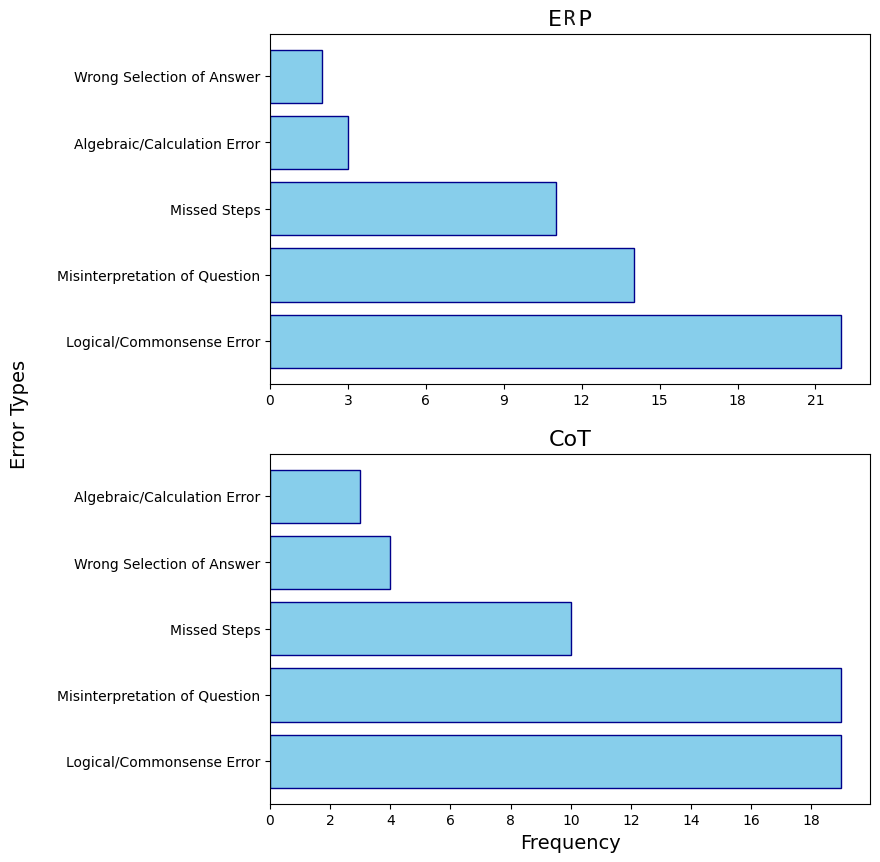} 
    \caption{Human annotations on GPT-4 errors in GSM8K.}
\label{fig:4}
\end{figure}


The correlation between error complexity and problem difficulty should be further researched to understand whether or not ERP may be effective with more complicated errors. Error outlining may not have been effective at reflection due to the discrepancy in difficulty and complexity.

\subsection{Commonsense Results}
\textbf{ERP scores higher on commonsense data versus CoT.}\indent GPT-4 + ERP had a slight increase over conventional CoT in CSQA, outperforming it by \textbf{+1.0\%} while GPT-3.5 + ERP had an increase of \textbf{+1.5\%}. For commonsense reasoning, ERP appears less effective than arithmetic reasoning. However, experimentation with the types of errors presented in the prompt may aid in reasoning.

\section{Error Analysis}
In this section, we explore how ERP affects the distribution of errors in answering questions. Although ERP scored higher than CoT \cite{wei2023chainofthought} for all the tests, the error distribution was not the same. This discrepancy is analyzed to explain ERP's behavior in a LLM. Confidence values for annotations were calculated by taking a subset of fifty randomly sampled annotated errors from both GPT-3.5 and GPT-4 inference errors and determining their validity. These values can be found in Table \ref{tab:5}. Let $T$ = GPT-3.5 errors and let $F$ = GPT-4 errors where both $T = \{0, 1\}$ and $F = \{0, 1\}$: 

\begin{center}
    $\frac{1}{100} \left( \sum_{i=1}^{50} \text{T}_i + \sum_{i=1}^{50} \text{F}_i \right)$
\end{center}

\subsection{Arithmetic Errors}
Figure \ref{fig:3} shows a set of errors from the annotations on math-based datasets. In each instance, ERP is less prone to Algebraic/Calculation Errors versus other types of errors. It is important to highlight that the prompts used did not contain instances of these errors, yet ERP exhibited a reduced occurence of such mistakes. However, it is also notable that the frequencies for selecting the wrong answer had also increased in a majority of cases. Figure \ref{fig:4} shows a full set of human annotated errors on GPT-4 on GSM8K \cite{DBLP:journals/corr/abs-2110-14168}. Indeed, the degree of effectiveness in targeting specific errors depends on the complexity of the prompt and the model's ability to comprehend the complexity of the error and its implications. For example, Misinterpretation errors were decreased in ERP versus CoT, although Missed Steps were less affected.

\subsection{Commonsense Errors}
Figure \ref{fig:5} illustrates the errors from commonsense reasoning datasets. Interestingly, ERP's utilization of error categories yielded mixed results compared to ERP without error categories and CoT \cite{wei2023chainofthought}. Despite its performance, ERP with error categories exhibited a notable capability in balancing error distribution, particularly evident in the case of GPT-3.5 in StrategyQA \cite{geva2021did}. This observation suggests that the incorporation of error categories within the ERP framework offers potential benefits in error reduction in certain categories, albeit with some trade-offs in overall performance metrics. Effectively, ERP without error categories highlights its potential utility in specific contexts where error reduction is paramount.

\section{Conclusion}
By exploring how weighting errors in explanations can enhance language models’ reasoning, we tackled some limitations of CoT. We found that highlighting and explaining common mistakes leads to more accurate, coherent responses. Though drawbacks remain—like overfitting to specific errors or creating effective rationales—our results show ERP’s versatility as a CoT extension: it not only identifies errors but also uncovers the steps that cause them. Future research into error-based prompting may further boost reasoning abilities and bring models closer to human-like thought processes.

\section{Limitations}

Although ERP shows promising results, it suffers from overfitting \cite{schmidt2023testing} to a particular dataset. The underlying issue lies in the prompt, where each prompt has to be tailored towards a specific task such as commonsense or arithmetic based reasoning sub-tasks.

Furthermore, there exists diminishing returns throughout the analysis of the results. ERP appears to be less effective as the model's number of parameters scales up \cite{qiu2023learn}. Further tuning with ERP, such as including more sophisticated and complicated set of errors, could allow a larger LLM to recognize more complex errors.

ERP is limited by the number of tokens it requires. Being a supplement to CoT, the prompt adds additional tokens that may increase costs, particularly when used in a few-shot setting. It's important to recognize the cost-to-accuracy ratio and whether or not ERP is sustainable in creating more accurate reasoning paths.

\bibliography{anthology,custom}

\newpage
\clearpage
\appendix
\section{Appendix A}
\label{app:Appendix A}

\small
\begin{center}
\begin{tabularx}{\textwidth}{XXX}
\textbf{PROMPTS FOR ERP -- GSM8K and MATH} \\
Input & Model Output \\
\hline

\\
\textbf{Example 1} \\
\underline{Question:} There are 15 trees in the grove. Grove workers will plant trees in the grove today. After they are done, there will be 21 trees. How many trees did the grove workers plant today? \newline \underline{Incorrect A:} There are initially 15 trees in the grove. To find the number of trees planted, we can subtract 21 from 15. 15 - 21 = -6. Since the answer is negative, the grove workers removed 6 trees from the grove, which means they planted 0. The answer is 0. & \underline{Errors:}
1. Incorrect Method - The wrong calculation is used. Subtracting the final number of trees from the initial number of trees is incorrect. The correct method is to subtract the initial number of trees from the final number to find how many were added. 
\underline{Correct A:} There are 15 trees originally. Then there were 21 trees after some more were planted. So there must have been 21 - 15 = 6. The answer is 6.\\

\\
\textbf{Example 2} \\
\underline{Question:} There were nine computers in the server room. Five more computers were installed each day, from Monday to Thursday. How many computers are now in the server room? \newline \underline{Incorrect A:} There were 9 computers in the server room. Five more computers are added each day, from Monday to Thursday. To find the number of computers in the server room, we can add 9 and 5. 9 + 5 = 14. Therefore, there are 14 computers in the server room. The answer is 14. & \underline{Errors:} 1. Missed step - The incorrect answer only added the computers from one day, not the computers added across four days, which leads to an incorrect total. \newline \underline{Correct A:} There were originally 9 computers. For each of 4 days, 5 more computers were added. So 5 * 4 = 20 computers were added. 9 + 20 is 29. The answer is 29.\\

\\
\textbf{Example 3} \\
\underline{Question:} Shawn has five toys. For Christmas, he got two toys each from his mom and dad. How many toys does he have now? \newline \underline{Incorrect A:} Shawn has 5 toys. He got 2 toys from his mom and dad, so he has 5 + 2 = 7 toys now. The answer is 7. & \underline{Errors:} 1. Misinterpretation Error - The incorrect answer adds only 2 in total, mistakenly combining the toys from both parents into a single gift. Shawn received 2 toys from his mom and 2 from his dad. \newline \underline{Correct A:} Shawn started with 5 toys. If he got 2 toys each from his mom and dad, then that is 4 more toys. 5 + 4 = 9. The answer is 9. \\

\\
\textbf{Example 4} \\
\underline{Question:} Michael had 58 golf balls. On Tuesday, he lost 23 golf balls. On Wednesday, he lost 2 more. How many golf balls did he have at the end of Wednesday? \newline \underline{Incorrect A:} Michael had 58 golf balls. He lost 23 golf balls on Tuesday, so he has 58 - 23 = 35 golf balls left. The answer is 35 golf balls. & \underline{Errors:} 1. Missed Step - The answer fails to account for the additional golf balls Michael lost on Wednesday. 2. Incorrect Final Calculation - By not including the golf balls lost on Wednesday, the final count of golf balls is incorrectly determined. \newline \underline{Correct A:} Michael started with 58 golf balls. After losing 23 on Tuesday, he had 58 - 23 = 35. After losing 2 more, he had 35 - 2 = 33 golf balls. The answer is 33. \\

\end{tabularx}
\end{center}
\clearpage

\begin{center}
\small
\begin{tabularx}{\textwidth}{XXX}
\textbf{PROMPTS FOR ERP -- AQuA} \\
Input & Model Output \\
\hline

\\
\textbf{Example 1} \\
\underline{Question:} Q: John found that the average of 15 numbers is 40. If 10 is added to each number then the mean of the numbers is? Answer Choices: (a) 50 (b) 45 (c) 65 (d) 78 (e) 64 \newline \underline{Incorrect A:} Adding 10 to each number will also add 10 to the average, making it 50, which is the new average of the numbers. Therefore, the answer is (c) 65. & \underline{Errors:}
Errors: 1. Wrong Selection of Answer - The incorrect answer incorrectly identifies choice (c) 65 as the correct answer when the right answer should have been choice (a) 50 following the logic given. \underline{Correct A:} If 10 is added to each number, then the mean of the numbers also increases by 10. So the new mean would be 50. The answer is (a).\\

\\
\textbf{Example 2} \\
\underline{Question:} If a / b = 3/4 and 8a + 5b = 22, then find the value of a. Answer Choices: (a) 1/2 (b) 3/2 (c) 5/2 (d) 4/2 (e) 7/2 \newline \underline{Incorrect A:}  To find the value of 'a', we use the given ratio a / b = 3/4 to express 'b' in terms of 'a'. We have b = (3a/4). Substitute this value into the equation 8a + 5b = 22: 8a + 5(3a/4) = 22. 8a + 15a/4 = 22. (32a + 15a) / 4 = 22. 47a / 4 = 22. 47a = 88. a = 88 / 47. a = 1.87234. Therefore, since the correct answer is not listed, there is no answer. & \underline{Errors:} 1. Arithmetic/Calculation Error - Misused the ratio to derive 'b' from 'a'. Expressing 'b' in terms of 'a' should have been b = (4a/3), not b = (3a/4). \newline \underline{Correct A:} If a / b = 3/4, then b = 4a / 3. So 8a + 5(4a / 3) = 22. This simplifies to 8a + 20a / 3 = 22, which means 44a / 3 = 22. So a is equal to 3/2. The answer is (b).\\

\\
\textbf{Example 3} \\
\underline{Question:} A person is traveling at 20 km/hr and reached his destiny in 2.5 hr then find the distance?
Answer Choices: (a) 53 km (b) 55 km (c) 52 km (d) 60 km (e) 50 km \newline \underline{Incorrect A:} The person traveled 20 km/hr for 2.5 hr. Therefore, he traveled 20 / 2.5 = 8 km. None of the answer choices are 8 km, so the answer is none of the above. & \underline{Errors:} 1. Incorrect Method Used - The incorrect answer divides speed by time, which is not the correct operation for finding distance. Distance should be calculated by multiplying speed by time. 2. Logical/Commonsense Error - The concept that speed divided by time would yield distance is incorrect. Distance is the product of speed and time. \newline \underline{Correct A:} The distance that the person traveled would have been 20 km/hr * 2.5 hrs = 50 km. The answer is (e). \\

\textbf{Example 4} \\
\underline{Question:} How many keystrokes are needed to type the numbers from 1 to 500?
Answer Choices: (a) 1156 (b) 1392 (c) 1480 (d) 1562 (e) 1788 \newline \underline{Incorrect A:} To determine the number of keystrokes needed to type the numbers from 1 to 500, we need to consider the number of digits in each number.
1. Count the number of single-digit numbers: There are 9 single-digit numbers (1 to 9), and each requires 1 keystroke.
2. Count the number of two-digit numbers: There are 90 two-digit numbers (10 to 99), and each requires 2 keystrokes.
3. Count the number of three-digit numbers: There are 400 three-digit numbers (100 to 500), and each requires 3 keystrokes.
Now, we can calculate the total number of keystrokes needed to type the numbers from 1 to 500: 9 * 1 + 90 * 2 + 400 * 3 = 1389.
The correct answer is not among the provided choices. The closest option is (b) 1392, but the correct answer is 1389 keystrokes. & \underline{Errors:} 1. Logical/Commonsense Error  - The count of three-digit numbers is incorrect. There are 401 three-digit numbers from 100 to 500, not 400. This is because 500 is also a three-digit number, and it is included in the range. \newline \underline{Correct A:} There are 9 one-digit numbers from 1 to 9. There are 90 two-digit numbers from 10 to 99. There are 401 three-digit numbers from 100 to 500. 9 + 90(2) + 401(3) = 1392. The answer is (b). \\
\end{tabularx}
\end{center}
\clearpage

\begin{table*}[t]
\centering
\small
\begin{tabularx}{\textwidth}{XXX}
\multicolumn{2}{l}{\textbf{PROMPTS FOR ERP -- StrategyQA (With Error Categories)}} \\
Input & Model Output \\
\hline
\\
\textbf{Example 1} \\
\underline{Question:} Do hamsters provide food for any animals? \newline \underline{Incorrect A:} No, hamsters do not provide food for any animals because they are often kept as pets and are protected by humans. They are too small to be considered an adequate food source for any larger animal and therefore are not part of the natural food chain. & \underline{Errors:} 1. Commonsense Error - The reasoning that hamsters do not provide food for any animals because they are kept as pets is flawed. Being a pet does not remove an animal from the food chain.
\underline{Correct A:} Hamsters are prey animals. Prey are food for predators. Thus, hamsters provide food for some animals. So the answer is yes. \\

\\
\textbf{Example 2} \\
\underline{Question:} Could Brooke Shields succeed at University of Pennsylvania? \newline \underline{Incorrect A:} Brooke Shields would not be able to succeed at the University of Pennsylvania because her acting skills have no relevance to academic success. The University of Pennsylvania values only mathematical and scientific acumen, and given that Brooke Shields is prominent in the arts, she would not possess the necessary skills to thrive in such an environment. Furthermore, success at one Ivy League institution does not translate to another, as each university has a completely different set of criteria for success. & \underline{Errors:} 1. Assumption Error - The assertion that Brooke Shields' acting skills have no relevance to academic success is incorrect, as many universities, including the University of Pennsylvania, have diverse programs where skills from various disciplines, including the arts, are valued. \newline \underline{Correct A:} Brooke Shields went to Princeton University. Princeton University is about as academically rigorous as the University of Pennsylvania. Thus, Brooke Shields could also succeed at the University of Pennsylvania. So the answer is yes. \\

\\
\textbf{Example 3} \\
\underline{Question:} Yes or no: Hydrogen’s atomic number squared exceeds number of Spice Girls? \newline \underline{Incorrect A:} Yes, hydrogen's atomic number squared exceeds the number of Spice Girls because when you square hydrogen's atomic number, which is a gas element, it expands due to the general properties of gases. As the atomic number expands, it becomes larger than solid numbers, such as the number of pERPle in a group like the Spice Girls. Since gases can fill their containers, the squared atomic number can be assumed to fill a container larger than the number 5, which represents the Spice Girls. & \underline{Errors:} 1. Commonsense Error - The answer is based on a misunderstanding of the scientific concepts involved. Atomic numbers do not expand like gases; they are constant and represent the number of protons in the nucleus of an atom.
\newline \underline{Correct A:} Hydrogen has an atomic number of 1. 1 squared is 1. There are 5 Spice Girls. Thus, Hydrogen’s atomic number squared is less than 5. So the answer is no. \\

\\
\textbf{Example 4} \\
\underline{Question:} Yes or no: Is it common to see frost during some college commencements? \newline \underline{Incorrect A:} No, frost is not common during college commencements because commencements are celebrations, and frost typically does not appear during festive events. PERPle usually associate celebrations with warmth and joy, so frost would not be present at such occasions. Additionally, commencements are often held indoors, so even if it were cold outside, the frost wouldn't affect the ceremonies. & \underline{Errors:} 1. Encyclopedic Error - We know that the statement \"frost typically does not appear during festive events\" is incorrect. Frost can occur during any event, regardless of its nature. \newline \underline{Correct A:} College commencement ceremonies can happen in December, May, and June. December is in the winter, so there can be frost. Thus, there could be frost at some commencements. So the answer is yes. \\
\end{tabularx}
\end{table*}
\clearpage

\begin{table*}
\begin{center}
\scriptsize
\begin{tabularx}{\textwidth}{XXX}
\multicolumn{2}{l}{\textbf{PROMPTS FOR ERP -- StrategyQA (No Error Categories)}} \\
Input & Model Output \\
\hline
\\
\textbf{Example 1} \\
\underline{Question:} Do hamsters provide food for any animals? \newline \underline{Incorrect A:} No, hamsters do not provide food for any animals because they are often kept as pets and are protected by humans. They are too small to be considered an adequate food source for any larger animal and therefore are not part of the natural food chain. & \underline{Errors:}
1. The reasoning that hamsters do not provide food for any animals because they are kept as pets is flawed. Being a pet does not remove an animal from the food chain; it only means that specific individuals are cared for by humans.
2. The claim that hamsters are too small to be considered an adequate food source is incorrect. Many predators in the wild prey on small mammals, including hamsters, such as snakes, birds of prey, and larger mammals.
3. The assertion that hamsters are not part of the natural food chain is false. Regardless of some being domesticated, as a species, hamsters are naturally part of various ecosystems and thus have predators that would feed on them.
4. The overall logic ignores the existence of wild hamsters and their role in the ecosystem, focusing only on domesticated hamsters.
\underline{Correct A:} Hamsters are prey animals. Prey are food for predators. Thus, hamsters provide food for some animals. So the answer is yes. \\

\\
\textbf{Example 2} \\
\underline{Question:} Could Brooke Shields succeed at University of Pennsylvania? \newline \underline{Incorrect A:} Brooke Shields would not be able to succeed at the University of Pennsylvania because her acting skills have no relevance to academic success. The University of Pennsylvania values only mathematical and scientific acumen, and given that Brooke Shields is prominent in the arts, she would not possess the necessary skills to thrive in such an environment. Furthermore, success at one Ivy League institution does not translate to another, as each university has a completely different set of criteria for success. & \underline{Errors:}
1. The assertion that Brooke Shields' acting skills have no relevance to academic success is incorrect, as many universities, including the University of Pennsylvania, have diverse programs where skills from various disciplines, including the arts, are valued.
2. The claim that the University of Pennsylvania values only mathematical and scientific acumen is incorrect. As an Ivy League institution, it offers a range of programs in the humanities, social sciences, business, and more, not just STEM fields.
3. The idea that success at one Ivy League school does not translate to another due to completely different criteria for success is flawed. While each institution is unique, all Ivy League universities have high academic standards, and the skill set that allows a person to thrive in one such environment is likely to be beneficial in another.
4. It is not considered to analyze Brooke Shields' actual academic record, which demonstrates her capability to succeed in an academically rigorous environment, thus making it likely she could succeed at a similar institution.
5. The conclusion drawn from this incorrect logic is flawed, as Brooke Shields has demonstrated academic prowess by graduating from Princeton University, and it is reasonable to assume she could succeed at an institution with similar academic rigor, such as the University of Pennsylvania.\newline \underline{Correct A:} Brooke Shields went to Princeton University. Princeton University is about as academically rigorous as the University of Pennsylvania. Thus, Brooke Shields could also succeed at the University of Pennsylvania. So the answer is yes. \\

\\
\textbf{Example 3} \\
\underline{Question:} Yes or no: Hydrogen’s atomic number squared exceeds number of Spice Girls? \newline \underline{Incorrect A:} Yes, hydrogen's atomic number squared exceeds the number of Spice Girls because when you square hydrogen's atomic number, which is a gas element, it expands due to the general properties of gases. As the atomic number expands, it becomes larger than solid numbers, such as the number of pERPle in a group like the Spice Girls. Since gases can fill their containers, the squared atomic number can be assumed to fill a container larger than the number 5, which represents the Spice Girls. & \underline{Errors:} 1. The answer is based on a misunderstanding of the scientific concepts involved. Atomic numbers do not expand like gases; they are constant and represent the number of protons in the nucleus of an atom.
2. Squaring an atomic number is a mathematical operation and has nothing to do with the physical properties of the element itself.
3. The analogy between the expansion of gases and the mathematical operation of squaring a number is incorrect. Mathematical operations do not behave like physical substances.
4. The implication that a 'solid number' is somehow fundamentally different from other numbers in this context is false. Numbers are abstract concepts and do not have physical states like solid, liquid, or gas.
\newline \underline{Correct A:} Hydrogen has an atomic number of 1. 1 squared is 1. There are 5 Spice Girls. Thus, Hydrogen’s atomic number squared is less than 5. So the answer is no. \\

\\
\textbf{Example 4} \\
\underline{Question:} Yes or no: Is it common to see frost during some college commencements? \newline \underline{Incorrect A:} No, frost is not common during college commencements because commencements are celebrations, and frost typically does not appear during festive events. PERPle usually associate celebrations with warmth and joy, so frost would not be present at such occasions. Additionally, commencements are often held indoors, so even if it were cold outside, the frost wouldn't affect the ceremonies. & \underline{Errors:} 1. The occurrence of frost is a weather phenomenon and has nothing to do with human activities or celebrations.
2. Associating warmth and joy with celebrations does not impact the actual weather conditions during the time of a commencement ceremony.
3. The answer incorrectly assumes that all commencement ceremonies are held indoors. While many are held indoors, some are held outside, and even if the ceremony is indoors, frost can still be common outside during the event in certain climates.
4. The incorrect answer overlooks the fact that college commencements in some regions may occur during colder months like December when frost is indeed a common occurrence.
5. The answer fails to consider global differences in weather patterns and seasons, which might lead to the presence of frost during commencements in different parts of the world.
\newline \underline{Correct A:} College commencement ceremonies can happen in December, May, and June. December is in the winter, so there can be frost. Thus, there could be frost at some commencements. So the answer is yes. \\
\end{tabularx}
\end{center}
\end{table*}

\begin{table*}
\begin{center}
\small
\begin{tabularx}{\textwidth}{XXX}
\multicolumn{2}{l}{\textbf{PROMPTS FOR ERP -- CSQA}} \\
Input & Model Output \\
\hline
\\
\textbf{Example 1} \\
\underline{Question:} What do pERPle use to absorb extra ink from a fountain pen? Answer Choices: (a) shirt pocket (b) calligrapher’s hand (c) inkwell (d) desk drawer (e) blotter \newline \underline{Incorrect A:} A desk drawer is used to store things, so it can be used to absorb extra ink. So the answer is (d). & \underline{Errors:}
1. Commonsense Error - The answer incorrectly assumes that a desk drawer can absorb ink, which is not possible. The answer should be an item that can absorb ink, not store it.
\underline{Correct A:} The answer must be an item that can absorb ink. Of the above choices, only blotters are used to absorb ink. So the answer is (e). \\

\\
\textbf{Example 2} \\
\underline{Question:} What home entertainment equipment requires cable? Answer Choices: (a) radio shack (b) substation (c) television (d) cabinet \newline \underline{Incorrect A:} A substation is a type of home entertainment equipment that requires cable. So the answer is (b). & \underline{Errors:}
1. Encyclopedic Error - The answer incorrectly identifies a substation as a type of home entertainment equipment that requires cable. A substation is not a home entertainment equipment. \underline{Correct A:} The answer must require cable. Of the above choices, only television requires cable. So the answer is (c). \\

\\
\textbf{Example 3} \\
\underline{Question:} The fox walked from the city into the forest, what was it looking for? Answer Choices: (a) pretty flowers (b) hen house (c) natural habitat (d) storybook \newline \underline{Incorrect A:} A fox may be looking for chickens to eat, so it was looking for a hen house. So the answer is (b). & \underline{Errors:}
1. Assumption Error - The answer incorrectly assumes that the fox was looking for chickens to eat. The question does not imply that the fox was looking for food. \underline{Correct A:} The answer must be something in the forest. Of the above choices, only natural habitat is in the forest. So the answer is (c). \\

\\
\textbf{Example 4} \\
\underline{Question:} Sammy wanted to go to where the pERPle were. Where might he go? Answer Choices: (a) populated areas (b) race track (c) desert (d) apartment (e) roadblock \newline \underline{Incorrect A:} The answer must be a place with a lot of pERPle, so Sammy would go to the desert because it has a lot of pERPle. So the answer is (c). & \underline{Errors:}
1. Self-Contradiction Error - The answer contradicts itself by stating that Sammy would go to the desert because it has a lot of pERPle. Deserts typically have very few pERPle, not a lot.
2. Assumption Error - The answer incorrectly assumes that the desert has a lot of pERPle, which is not true. \underline{Correct A:} The answer must be a place with a lot of pERPle. Of the above choices, only populated areas have a lot of pERPle. So the answer is (a). \\
\end{tabularx}
\end{center}
\end{table*}

\clearpage
\section{Error Types Table}
\label{app:error-types-table}

\begin{table*}[ht]
  \centering
  \renewcommand{\arraystretch}{1}
  \label{tab:error-types}
  \begin{tabularx}{\textwidth}{XXX} 
    \hline
    \toprule
    Error Type & Definition & Example \\
    \hline
    \midrule
    \textbf{Arithmetic Errors} \\
    \ \ Misinterpretation & Misinterpretation of the question's information or tasks& Prompt: Timothy has \textbf{\( \frac{1}{4} \) less} cookies... Model: Timothy has \textbf{\( \frac{1}{4} \) as many}... \\
    \midrule
    \ \ Missed Steps & Missing an extra step to completely solve the problem& Prompt: ...leave your answer in \textbf{hours} Model: ...answer is 360 \textbf{minutes}.\\
    \midrule
    \ \ Logical/Commonsense & Logic does not follow natural laws of the world& To solve this, we use $d$ = \( \frac{r}{t} \)...\\
    \midrule
    \ \ Wrong Selection & The presented answer is different from the calculated one& ...comes out to 56, which corresponds to option \textbf{(A)}. Therefore, the answer is \textbf{(B)}.\\
    \midrule
    \ \ Algebraic/Calculation & Arithmetic operations are wrong. This includes operations in algebraic equations& 102 + 56 = \textbf{160}\\
    \hline
    \midrule
    \textbf{Commonsense Errors} \\
    \ \ Self-Contradiction & Answer's reasoning is contradictory& ...a very bright environment, so the dark would be suitable... \\
    \midrule
    \ \ Assumption & Assuming information that is not given in the question& Assuming all animals like hot environments...\\
    \midrule
    \ \ Encyclopedic & Factual information that the annotator knows is wrong& Wallets are a specialized type of water bottle\\
    \midrule
    \ \ Commonsense & The answer violates commonsense and basic understanding of natural laws& ...the sun is as cold as ice cream.\\
    \hline
    \bottomrule
  \end{tabularx}
  \caption{Error types used in analysis.}
  \label{tab:2}
\end{table*}

\clearpage
\section{Annotation Confidence Values}
\label{app:annotation-confidence}

\begin{table}[H]
    \centering
    \small
    \resizebox{\linewidth}{!}{%
    \begin{tabular}{@{}lcccc@{}}
    \toprule
         & GSM8K & AQuA & StrategyQA & CSQA \\ 
    \midrule
         CoT & 0.60 & 0.76 & 0.75 & 0.79 \\
         ERP & 0.63 & 0.63 & 0.70 & 0.80 \\
         ERP Classes & -- & 0.69 & -- & -- \\
    \bottomrule
    \end{tabular}%
    }
    \caption{Confidence values for GPT-4 annotation.}
    \label{tab:5}
\end{table}

\section{Additional Error Annotations}
\label{app:commonsense-figure}

\begin{figure*}[h]
    \begin{center}
        \includegraphics[width=0.8\textwidth]{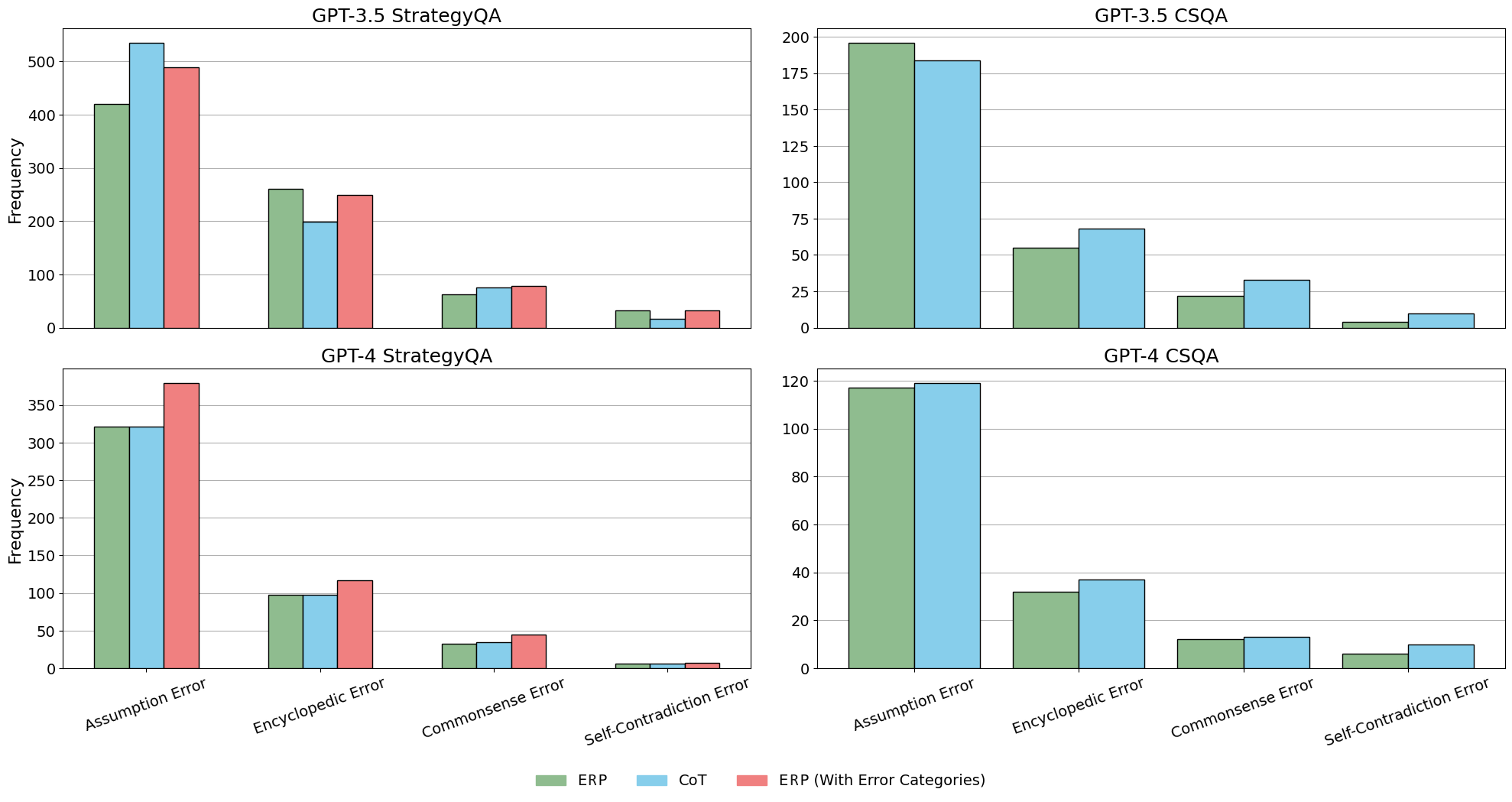}
    \end{center}
    \caption{Error annotations on arithmetic reasoning datasets.}
    \label{fig:5}
\end{figure*}

\end{document}